\documentclass[runningheads]{llncs}
\usepackage{graphicx}
\usepackage{comment}
\usepackage{amsmath,amssymb} 
\usepackage{color}
\usepackage{multirow}
\usepackage{enumitem}
\usepackage{mmstyle}
\usepackage{breakcites}
\usepackage[colorlinks,linkcolor=blue]{hyperref}

\usepackage{microtype}

\begin{document}
\pagestyle{headings}
\mainmatter
\def\ECCVSubNumber{3644}  

\title{Placepedia: Comprehensive Place Understanding with Multi-Faceted Annotations} 


\titlerunning{Placepedia: Comp. Place Understanding with Multi-Faceted Annotations}


\author{
	Huaiyi Huang\orcidID{0000-0003-1548-2498} \and
	Yuqi Zhang\orcidID{0000-0003-1883-4081} \and \\
	Qingqiu Huang\orcidID{0000-0002-6467-1634} \and
	Zhengkui Guo\orcidID{0000-0002-8093-7578} \and \\
	Ziwei Liu\orcidID{0000-0002-4220-5958} \and
	Dahua Lin\orcidID{0000-0002-8865-7896}
}

\authorrunning{H. Huang and Y. Zhang and Q. Huang and Z. Guo and Z. Liu and D. Lin}

\institute{The Chinese University of Hong Kong \\
\email{\{hh016,zy016,hq016,gz019,dhlin\}@ie.cuhk.edu.hk}, zwliu.hust@gmail.com}

\maketitle
\begin{figure}
    \centering
    \includegraphics[width=0.9\linewidth]{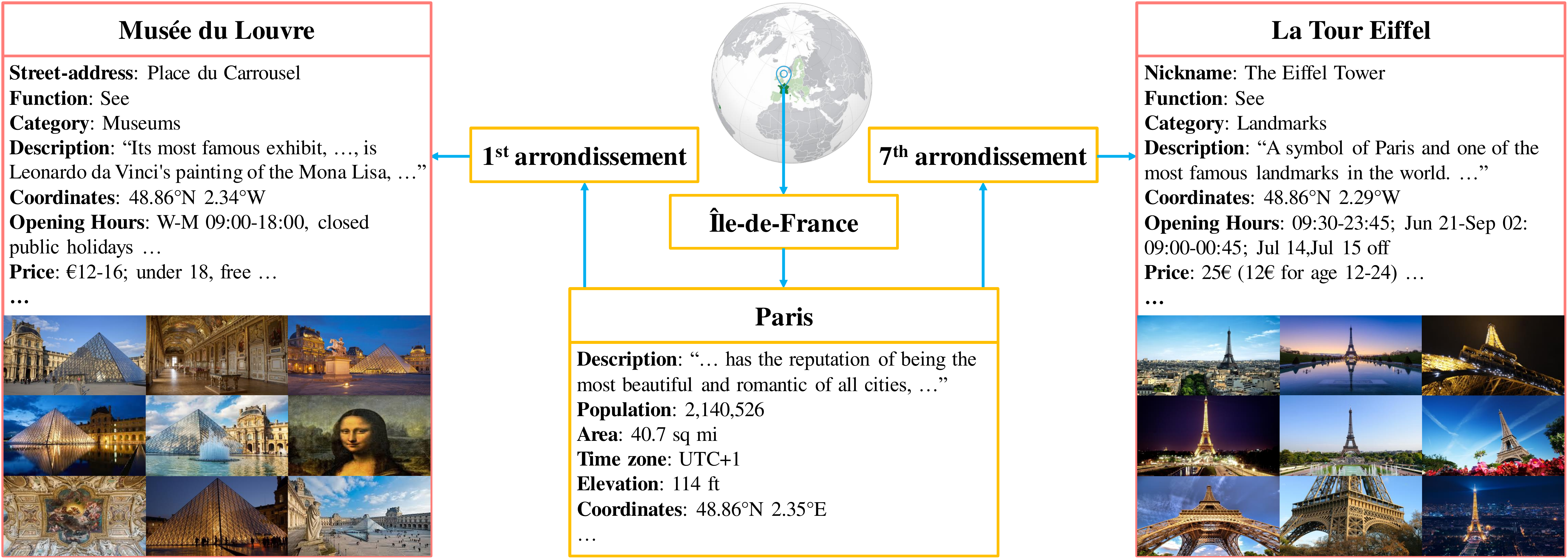}
    \caption{Hierarchical structure of Placepedia with places from all over the 
    world. Each place is associated with its \emph{district}, 
    \emph{city/town/village}, \emph{state/province}, \emph{country}, 
    \emph{continent}, and a large amount of diverse photos. Both administrative 
    areas and places have rich side information, \eg \emph{description}, 
    \emph{population}, \emph{category}, \emph{function}, which allows various 
    large-scale studies to be conducted on top of it}
\label{fig_teaser}
\end{figure}%


\begin{abstract}
Place is an important element in visual understanding. 
Given a photo of a building, people can often tell its functionality, \eg~a restaurant or a shop, its cultural style, \eg~Asian or European, as well as its economic type, \eg~industry oriented or tourism oriented.
While place recognition has been widely studied in previous work, there remains a long way towards comprehensive place understanding, which is far beyond categorizing a place with an image and requires information of multiple aspects.
In this work, we contribute Placepedia\footnote{The dataset is available at: \href{https://hahehi.github.io/placepedia.html}{https://hahehi.github.io/placepedia.html}}, a large-scale place dataset with more than $35M$ photos from $240K$ unique places.
Besides the photos, each place also comes with massive multi-faceted information, \eg GDP, population, \etc, and labels at multiple levels, including function, city, country, \etc.
This dataset, with its large amount of data and rich annotations, allows various studies to be conducted.
Particularly, in our studies, we develop 1) PlaceNet, a unified framework for multi-level place recognition, and 2) a method for city embedding, which can produce a vector representation for a city that captures both visual and multi-faceted side information.
Such studies not only reveal key challenges in place understanding, but also establish connections between visual observations and underlying socioeconomic/cultural implications.

\end{abstract}

\section{Introduction}

Imagine that you are visiting a new country, and you are traveling among different cities. In each city, you will encounter countless places, and you may see a fancy building, experience some natural wild beauty, 
or enjoy the unique culture, \etc.
All of these experiences impress you and lead you to a deeper understanding of the place.
As we browse through a city, based on certain common visual elements therein, we implicitly establish connections between 
visual characteristics with other multi-faceted information, such as its function, socioeconomic status and culture. 
We therefore believe that it would be a rewarding adventure to 
move beyond conventional place categorization and 
explore the connections among different aspects of a place. 
It indicates that multi-dimension labels are essential for comprehensive place understanding.
To support this exploration, a large-scale dataset that cover a diverse set of 
places with both images and comprehensive multi-faceted information is needed.

However, existing datasets for place 
understanding~\cite{noh2017large,zhou2017places,philbin2008lost}, 
as shown in Tab.~\ref{tab:datasets_stat} are subject to at least one
of the following drawbacks:
1) \emph{Limited Scale.} Some of them \cite{philbin2007object,philbin2008lost} contain only several thousand images from one particular 
city.
2) \emph{Restrictive Scope.} Most datasets are constructed for only one task, \eg 
place retrieval \cite{noh2017large} or scene recognition 
\cite{zhou2017places,jegou2008hamming}.
3) \emph{Lack of Attributes.} These datasets often contain just a very limited set
of attributes. For example, \cite{noh2017large} contains just photographers and 
titles.
Clearly, these datasets, due to their limitations in scale, diversity, and richness, 
are not able to support the development of comprehensive place understanding.

In this work, we develop \emph{Placepedia},
a comprehensive place dataset that contains images for places of interest 
from all over the world with massive attributes,
as shown in Fig.~\ref{fig_teaser}.
\emph{Placepedia} is distinguished in several aspects:
{\bf 1)} \emph{Large Scale.}
It contains over $35M$ images from $240K$ places, several times larger than previous ones.
{\bf 2)} \emph{Comprehensive Annotations.}
The places in Placepedia are tagged with categories, functions, 
administrative divisions at different levels, \eg city and country, 
as well as lots of multi-faceted side information, \eg descriptions and coordinates. 
{\bf 3)} \emph{Public Availability.}
Placepedia will be made public to the research community.
We believe that it will greatly benefit the research on comprehensive place understanding 
and beyond.


\begin{table}
	\caption{Comparing Placepedia with other existing datasets. Placepedia offers the largest number of images and the richest information}
	\label{tab:datasets_stat}
\centering
\scriptsize
{
\begin{tabular}{c|c|c|c|c}
\hline
& \# places & \# images & \# categories & Meta data \\
\hline
\hline
Google Landmarks \cite{noh2017large}& 203,094  & 5,012,248 & N/A & authors, titles\\
\hline
Places365 \cite{zhou2017places}& N/A & 10,624,928 & 434 & N/A\\
\hline
Holidays \cite{jegou2008hamming} & N/A & 1,491 & 500 & N/A\\
\hline
Oxford 5k \cite{philbin2007object} & 11 & 5,062 & N/A & N/A\\
\hline
Paris 6k \cite{philbin2008lost} & 11 & 6,412 & N/A & N/A\\
\hline
SFLD \cite{chen2011city} & N/A & 1,700,000 & N/A & coordinates\\
\hline
Pitts 250k \cite{torii2013visual} & N/A & 254,064 & N/A & coordinates\\
\hline
Google Street View \cite{gronat2011building} & 10,343 & 62,058 & N/A & coordinates, addresses, \etc\\
\hline
Tokyo 24/7 \cite{torii201524} & 125 & 74,000 & N/A & coordinates\\
\hline
Cambridge Landmarks \cite{en2018rpnet} & 6 & \textgreater 12,000 & N/A & N/A\\
\hline
Vietnam Landscape$^a$ & 103 & 118,000 & N/A & N/A\\
\hline
\textbf{Placepedia} & \textbf{\textgreater 240,000} & \textbf{\textgreater 
35,000,000} & \textbf{\textgreater 3,000} & \textbf{\tiny{divisions, descriptions, city info, \etc}}\\
\hline
\end{tabular}
}
\footnotesize{$^a$ https://blog.facebit.net/2018/09/07/zalo-ai-challenge-problems-and-solutions}\\
\end{table}

Meanwhile, Placepedia also enables us to rigorously
benchmark the performance of existing and future algorithms
for place recognition.
We create four benchmarks based on Placepedia in this paper,
namely \emph{place retrieval}, 
\emph{place categorization}, \emph{function categorization}, 
and \emph{city/country recognition}.
By comparing different methods and modeling choices on these benchmarks, 
we gain insights into their pros and cons, which we believe would inspire 
more effective techniques for place recognition.
Furthermore, to provide a trigger for comprehensive place understanding, 
we develop \emph{PlaceNet}, a unified deep network for multi-level place recognition.
It simultaneously predicts place item, category, function, city, and country.
Experiments show that by leveraging the multi-level annotations in Placepedia, 
PlaceNet can learn better representation of a place than previous works.
We also leverage both visual and side information from 
Placepedia to learn city embeddings, which demonstrate strong expressive power 
as well as the insights on \emph{what distinguish a city}. 

From the empirical studies on Placepedia,
we see lots of challenges in performing place recognition.
1) The visual appearance can vary significantly due to the changes of angle, 
illumination, and other environmental factors.
2) A place may look completely different when viewed from inside and outside.
3) A big place, \eg~a university, usually consists of a number of small places that have nothing
in common in appearance.
All these problems remain open.
We hope that Placepedia, with its large scale, high diversity and massive annotations, 
would provide a gold mine for the community to develop more expressive models 
to meet the aforementioned challenges.

Our contributions in this work can be summarized as below.
{\bf 1)} We build Placepedia, a large-scale place dataset with comprehensive annotations 
in multiple aspects. To the best of our knowledge, Placepedia is the largest and 
the most comprehensive dataset for place understanding.
{\bf 2)} We design four task-specific benchmarks on Placepedia \wrt the multi-faceted information.
{\bf 3)} We conduct systematic studies on place recognition and city embedding,
which demonstrate important challenges in place understanding as well as 
the connections between the visual characteristics and the underlying socioeconomic or cultural implications.

\section{Related Work}

\noindent
\textbf{Place Understanding Datasets.}
Datasets play an important role for various research topics in computer vision~\cite{krizhevsky2012imagenet,liu2015deep,caba2015activitynet,liu2016deepfashion,loy2019wider,huang2020movie,rao2020unified,huang2020caption,rao2020local,Xiong_2019_ICCV,huang2018person,Huang_2018_CVPR,huang2018trailers,yang2020propagate,zhang2018accelerated,yang2019learning,yang2020learning}.
During the past decade, lots of place datasets were constructed to facilitate 
place-related studies. There are mainly three kinds of datasets.
The first kind \cite{noh2017large,torii201524,chen2011city,torii2013visual,gronat2011building} focuses on the tasks of place 
recognition/retrieval, where images are labeled as 
particular place items, e.g. White House or Eiffel Tower. The second kind 
\cite{zhou2017places} targets place categorization or scene recognition. In 
these datasets, each image is attached with a place type, e.g. parks or 
museums. The third kind \cite{philbin2007object,philbin2008lost,jegou2008hamming} is for object/image retrieval. The statistics is summarized 
in Tab. \ref{tab:datasets_stat}. Compared with these datasets, our Placepedia 
has much larger amount of image and context data, containing over 
$240K$ places with $35$ million images labeled with $3K$ categories. Hence, 
Placepedia can be used for all these tasks. Besides, the provision 
of hierarchical administrative divisions for places allows us to study 
place recognition in different scales, \eg city recognition or country recognition. Also, the 
function information (\emph{See}, \emph{Do}, \emph{Sleep}, \emph{Eat}, 
\emph{Buy}, etc.) of places may lead to a new task, namely place function recognition.

\noindent
\textbf{Place Understanding Tasks.}
Lots of work aims at 2D place recognition \cite{chen2011city,torii2013visual,gronat2011building,torii201524,arandjelovic2014dislocation,cao2013graph,gronat2013learning,knopp2010avoiding,sattler2015hyperpoints,schindler2007city,arandjelovic2016netvlad,stumm2016robust,lopez2017appearance,mishkin2015place,milford2015sequence,sizikova2016enhancing,panphattarasap2016visual,zhu2018attention,johns2015ransac,hong2019textplace} or place retrieval \cite{noh2017large,teichmann2019detect,johns2015ransac,sattler2012image,wang2015effective,gavves2012visual,gavves2010landmark,sun2008place}. Given an image, the goal is to recognize what the particular place is or to retrieve images representing the same place. Scene recognition \cite{zhou2017places,zhou2014learning,yang2015scene,li2009landmark,xiao2010sun,torralba200880}, on the other hand, defines a diverse list of environment types as labels, \eg nature, classroom, bar. And their job is to assign each image a scene type. \cite{doersch2012makes,shi2019deep} collects images from several different cities, studies on distinguishing images of one city from others, and discovers what elements are characteristics of a given city. 
There also exist some other humanities-related studies. \cite{son2018your} classifies keywords of city description into \emph{Economic}, \emph{Cultural}, or \emph{Political}, and then counts the occurrences of these three types to represent city branch. \cite{lu2018mapping} uses satellite data from both daytime and nighttime to discover ``ghost cities'' which consist of abandoned buildings or housing structures which may hurt urbanization process. \cite{kang2019extracting} collects place images from social media to extract human emotions at different places, to find out the relationship between human emotions and environment factors. \cite{nice2019paris} uses neural networks trained with map imagery to understand the connection among cities, transportation, and human health.
%
Placepedia, with its large amount of data in both visual and textual domains, 
allows various studies to be conducted on top in large scale.


\section{The Placepedia Dataset}
\begin{figure}[h]
\center
\includegraphics[width=0.9\linewidth]{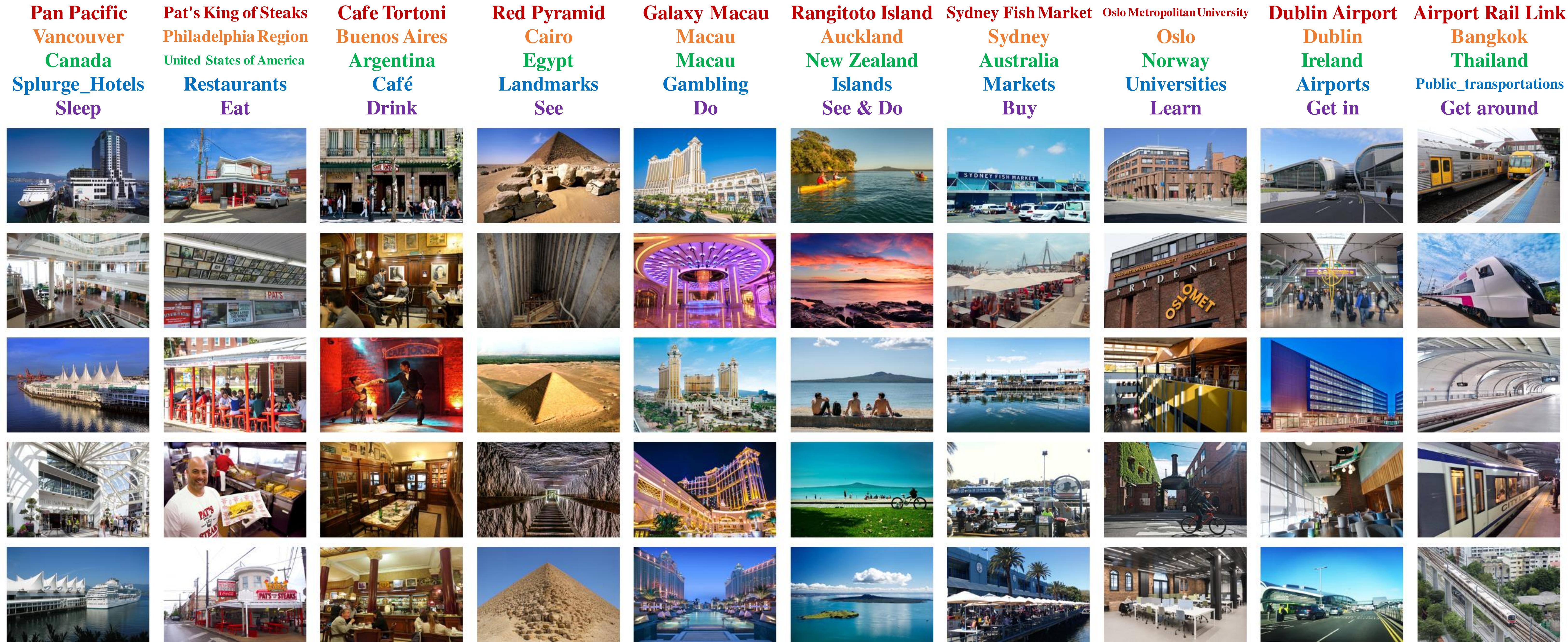}
\caption{The text in red, orange, green, blue, purple represents place names, cities, 
countries/territories, categories, functions, respectively. We see that the appearance of a place can vary from: 1) daytime to nighttime, 2) different angles, 3) inside and outside}
\label{fig_examples}
\end{figure}
\begin{figure}[h]
\center
\includegraphics[width=0.9\linewidth]{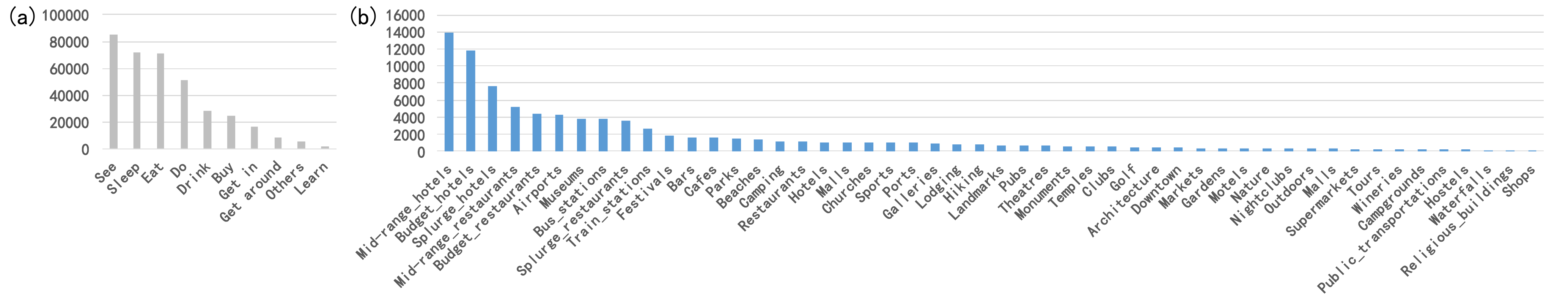}
\caption{(a) The number of places for top-$10$ functions. (b) The number of places for top-$50$ categories}
\label{fig_funccate}\end{figure}


We contribute Placepedia, a large-scale place dataset, to the community. Some 
example images along with labels are shown in Fig. \ref{fig_examples}. In 
this section, we introduce the procedure of building Placepedia. First, a hierarchical structure is organized to store places and their multi-level administrative divisions, where each place/division is associated with rich side information. With this structure, global places are connected and classified on different levels, which allows us to investigate numerous place-related issues, \eg city recognition \cite{doersch2012makes,shi2019deep}, country recognition, and city embedding. With types (\eg \emph{park}, \emph{airport}) provided we can explore tasks such as place categorization \cite{zhou2017places,li2009landmark}. With functions (\eg \emph{See}, \emph{Sleep}) provided we are able to model the functionality of places.
Second, we download place images from Google Image, which are cleaned automatically and manually.

\subsection{Hierarchical Administrative Areas and Places}
\label{hierarchical}
\noindent\textbf{Place Collection.} We collect place items with side information from 
Wikivoyage\footnote{https://en.wikivoyage.org/wiki/Destinations}, a free 
worldwide travel guide website, through the public channel. Pages of Wikivoyage are organized in a hierarchical way, \ie, each destination is obtained by walking through its \emph{continent}, \emph{country}, \emph{state}/\emph{province}, \emph{city}/\emph{town}/\emph{village}, \emph{district}, \etc. As illustrated in Fig. \ref{fig_teaser}, these administrative areas serve as non-leaf nodes in Placepedia,
and leaf nodes represent all the places.
This process results in a list of $361,524$ places together with $24,333$ 
administrative areas.

\begin{figure}[h]
	\center
	\includegraphics[width=0.9\linewidth]{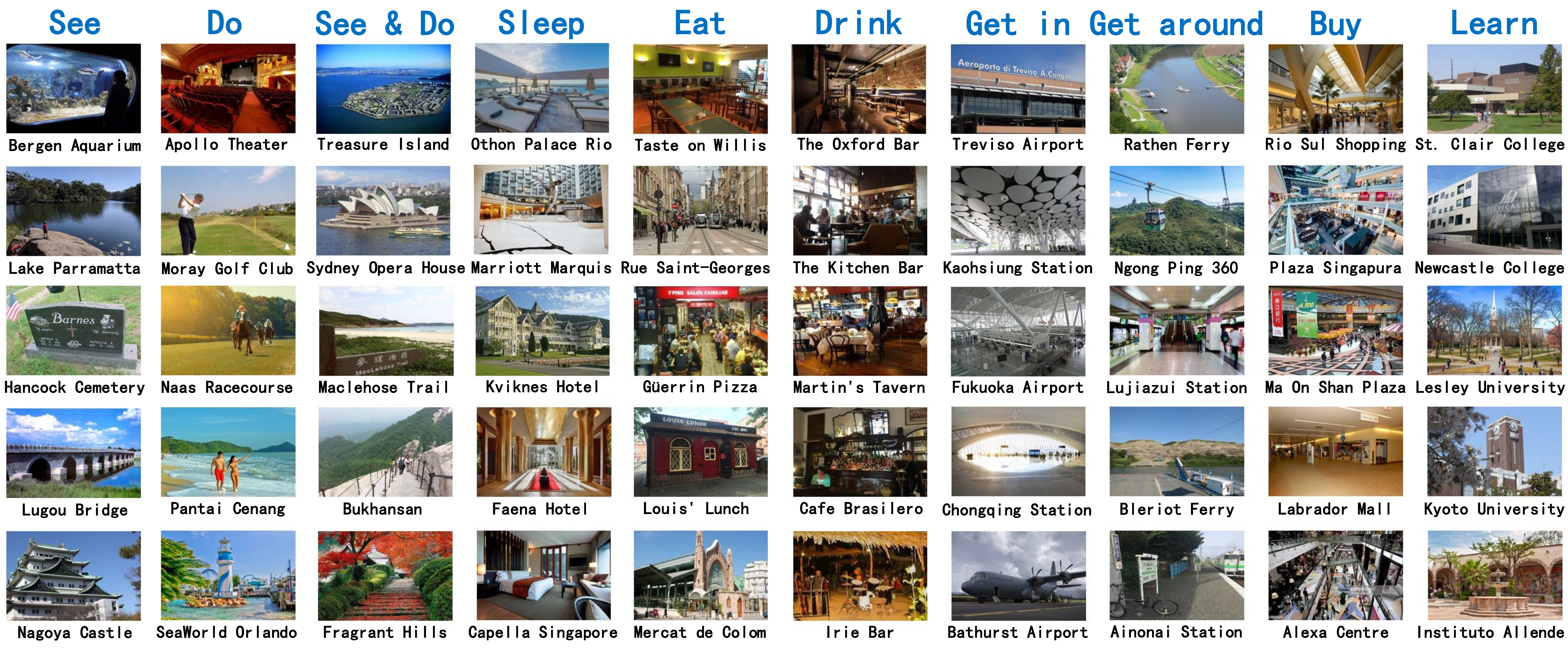}
	\caption{This figure shows ten \emph{function} labels with five example places for each}
	\label{fig_funcexamples}
\end{figure}
\begin{table}[]
    \centering
    \scriptsize
\caption{The description and examples for the 10 \emph{function} labels of Places-Fine and Places-Coarse, which are collected from Wikivoyage}
    \begin{tabular}{l|l|l|l}
    \hline
    Label & Description & Category examples & Place examples \\
    \hline
    \multirow{2}*{See} & People can enjoy beautiful scenes, arts, & Park, Tower, Museum,  & Bergen Aquarium,\\
    & and architectures therefrom. & Gallery, Historical\_site & Lake Parramatta\\
    \hline
    \multirow{3}*{Do} & People can do significant things such as  & Library, Theater, & Apollo Theater,\\
    & reading books, watching movies, going & Resort, Sport, Beach & Moray Golf Club\\
    & on vacation, playing sports. && \\
    \hline
    \multirow{3}*{See \& Do} & People can \emph{see} and \emph{do} in these places. & Land\_nature, Theater, & Treasure Island,\\
    & For instance, people can not only enjoy  & Resort, Park, Island & \tiny{Sydney Opera House}\\
    & mountain landscape but also climb them. && \\
    \hline
    \multirow{2}*{Sleep} & \multirow{2}*{People can have a sleep there.} & Splurge\_hotel, & Othon Palace Rio,\\
    & & Mid-range\_hotel & Kviknes Hotel\\
    \hline
    \multirow{2}*{Eat} & \multirow{2}*{People can eat food there.} & \multirow{2}*{Restaurant, Street} & Louis' Lunch,\\
    & & & Taste on Willis\\
    \hline
    \multirow{2}*{Drink} & \multirow{2}*{People can drink something there.} & \multirow{2}*{Café, Pub, Street} & The Oxford Bar,\\
    & & & Cafe Brasilero\\
    \hline
    \multirow{2}*{Get in} & Places for intercity or intercountry & Airport, Train\_station, & Treviso Airport,\\
    & transportation. & Public\_transportation & Kaohsiung Station\\
    \hline
    \multirow{3}*{Get around} & People can travel from one place to & Train\_station, & Lujiazui Station,\\
    & another inside a city or a town, such as & Public\_transportation & Rathen Ferry\\
    & bus stations, metros, and some ports. &&\\
    \hline
    \multirow{2}*{Buy} & \multirow{2}*{People usually go shopping there.} & Mall, Market, Street, & Labrador Mall,\\
    & & Shop, Town, Square& Alexa Center\\
    \hline
    \multirow{2}*{Learn} & People usually learn new knowledge or & \multirow{2}*{University, Sport} & Kyoto University,\\
    &  skills there. & & St. Clair College\\
    \hline
    \end{tabular}
    \label{tab:function}
\end{table}


\noindent\textbf{Meta Data Collection.} In Wikivoyage, all destinations are associated 
with some of the attributes below: \emph{function}, \emph{category}, 
\emph{description}, \emph{GPS coordinates}, \emph{address}, \emph{phone 
number}, \emph{opening hour}, \emph{price}, \emph{homepage}, \emph{wikipedia 
link}. 
The place number of top-$10$ functions 
and top-$50$ categories are shown in Fig. \ref{fig_funccate}. 
Tab. \ref{tab:function} shows the definition for ten functions in Placepedia and Fig. \ref{fig_funcexamples} shows several examples of these ten \emph{function}s. Function labels are the section names of places from Wikivoyage. Place functions serve as a good indicator for travelers to choose where to go. For example, some people love to go shopping when traveling; Some prefer to enjoy various flavors of food; Some people are addicted to distinctive landscapes.
For 
administrative areas, 
Wikivoyage often lack meta data. Hence, we acquire the missing information by 
parsing their Wikipedia page. At last, the following 
attributes are extracted: \emph{description}, \emph{GDP}, \emph{population 
density}, \emph{population}, 
\emph{elevation}, \emph{time zone}, \emph{area},
\emph{GPS coordinates}, \emph{establish time}, \etc.

\noindent\textbf{Place Cleaning.} To refine the place list, we only keep places 
satisfying at least one of the two following criteria: 
1) It has the attribute \emph{GPS coordinates} or \emph{address}; 
2) It is identified as a 
\emph{location} by Google Entity Recognition 
\footnote{https://cloud.google.com/natural-language} or Stanford 
Entity Recognition \footnote{https://nlp.stanford.edu/software/CRF-NER.html}. After the 
removal, $44,997$ items are deleted, and $316,527$ valid place entities remain.

\subsection{Place Images}
\label{placeimages}
\noindent\textbf{Image Collection.} We collect all place images from Google Image engine in the public domain.
For each location, its name plus its country is used as the keyword for searching.
To increase the probability that images are relevant to a particular location, we only download those whose stem words of image titles contain all stem words of the location name. By this process, a total of over $30M$ images are collected from Google Image.

\noindent\textbf{Image Cleaning.} There are $28,154$ places containing Wikipedia links with $8,125,108$ images. We use this subset to further study place-related 
tasks. Image set is refined by two stages. Firstly, we use Image Hashing to 
remove 
duplicate images.
Secondly, we 
ask human annotators to remove irrelevant images for each place, including those: 1) whose main body represents another place; 2) that are selfie images with faces occupying a large proportion; 3) that are maps indicating the geolocation of the place.
In total, $26K$ places with $5M$ images are kept to form this subset.
For those places without 
category labels, we manually annotate the labels for them. And after merging 
some similar labels, we obtain 50 categories.

Placepedia also helps solve some problems on place understanding, like label confusion and label noise. On one hand, all the labels are collected automatically from the Wikivoyage website. Since it is a popular website that provides worldwide travel guidance and the labels in Wikivoyage have been well organized, there are less label confusion in Placepedia dataset. On the other hand, we have manually checked the labels in Placepedia, which would significantly reduce label noise.


From the examples shown in 
Fig. \ref{fig_examples}, we observe that:
1) Images of places may look changeable from daytime to nighttime or during different seasons;
2) It can be significantly different viewed from multiple angles;
3) The appearances from inside and outside usually have little in common;
4) Some places such as universities span very large area and consist of different types of small places.
These factors make place-related tasks very challenging.
In the rest of this paper, we conduct a series of experiments to demonstrate important challenges in place understanding as well as strong expressive power of city embeddings.


\section{Study on Comprehensive Place Understanding}
\label{exp}

This section introduces our exploration on comprehensive place understanding.
Firstly, we carefully design benchmarks, and we evaluate the dataset with a lot of state-of-the-art models with different backbones for different tasks. Secondly, we develop a multi-task model, PlaceNet, which is trained to simultaneously predict place items, categories, functions, cities, and countries. This unified framework for place recognition can serve as a reasonable baseline for further studies in our Placepedia dataset. From the experimental results we also demonstrate the challenges of place recognition on multiple aspects.

\subsection{Benchmarks}
\label{benchmarks}

We build the following benchmarks based on the well-cleaned Placepedia subset, for evaluating different methods.

\paragraph{Datasets}
\setlist[itemize]{leftmargin=*}
\begin{itemize}
\item{\emph{Places-Coarse.}} We select $200$ places for validation and $400$ places for testing, from $50$ famous cities of $34$ countries.
The remained $25K$ places are used for training. For validation/testing set, we double checked the annotation results.
Places without category labels are manually annotated. After merging similar items of labels, we obtain $50$ categories and $10$ functions.
The training/validation/testing set have $5M$/$60K$/$120K$ images respectively, from $7K$ cities of more than $200$ countries.

\item{\emph{Places-Fine.}} Places-Fine shares the same validation/testing set with Places-Coarse.
For training set, we selected $400$ places from the $50$ cities of validation/testing places.
Different from Places-Coarse, we also double checked the annotation of training data. The training/validation/testing set have $110K$/$60K$/$120K$ images respectively, which are tagged with $50$ categories, $10$ functions, $50$ cities, and $34$ countries.
\end{itemize}

\paragraph{Tasks}
\begin{itemize}
\item{\emph{Place Retrieval.}} This task is to determine if two images belong to the 
same place. It is important when people want to find more photos of places 
they adore. For validation and testing set, $20$ images for each place are 
selected as queries and the rest images form the gallery. Top-$k$ retrieval 
accuracy is 
adopted to measure the performance of place retrieval, such that a successful
retrieval is counted if at least one image of the same place has been found
in the top-$k$ retrieved results.

\item{\emph{Place Categorization.}} This task is to classify places into $50$ place 
categories, 
e.g. \emph{museums}, \emph{parks}, \emph{churches}, \emph{temples}. 
For place categorization, we employ the standard top-$k$ classification accuracy as evaluation metric.

\item{\emph{Function Categorization.}} This task is to classify places into $10$ place
functions: \emph{See}, \emph{Do}, \emph{Sleep}, \emph{Eat}, \emph{Drink}, 
\emph{See \& Do}, \emph{Get In}, \emph{Get Around}, \emph{Buy}, 
\emph{Learn}.
Again, we employ the 
standard top-$k$ classification accuracy as 
evaluation metric.

\item{\emph{City/Country Recognition.}} This task is to classify places into $50$ cities 
or $34$ 
countries. The goal is to determine what city/country an image 
belongs to. Also, the standard top-$k$ classification accuracy is applied as 
evaluation metric.

\end{itemize}


\begin{figure}[t]
	\center
	\includegraphics[width=0.5\linewidth]{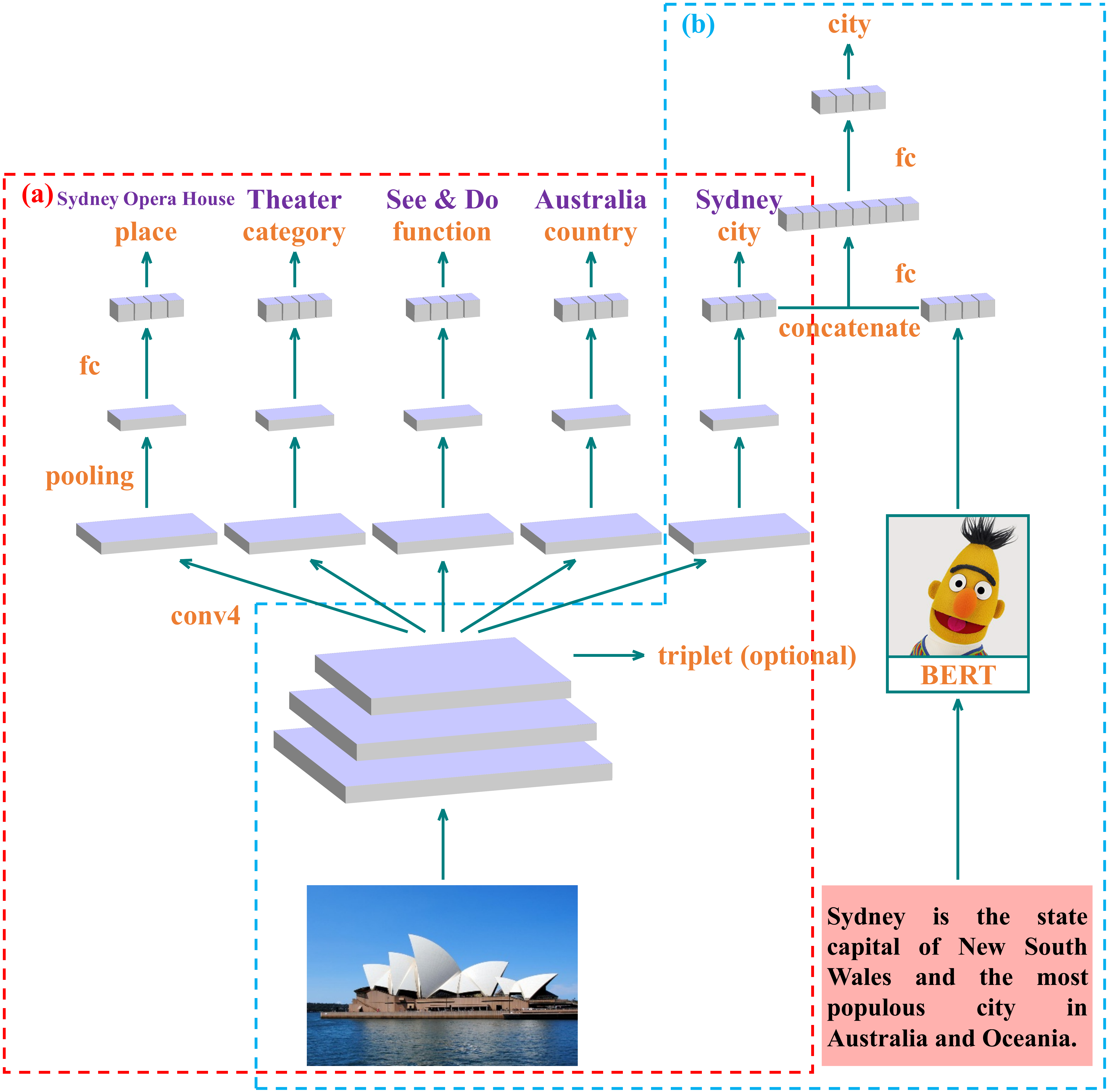}
	\caption{(a) Pipeline of PlaceNet, which learns five tasks simultaneously. 
	(b) Pipeline of city embedding, which learns city representations considering both vision and text information}
	\label{fig_networks}
\end{figure}

\subsection{PlaceNet}
We construct a CNN-based model to predict all tasks simultaneously. The 
training procedure performs in an iterative manner and the system is 
learned end-to-end.

\noindent\textbf{Network Structures.} The network structure of PlaceNet is similar to 
ResNet50 \cite{he2016deep}, which has been demonstrated powerful in various 
vision tasks. As illustrated in Figure \ref{fig_networks} (a), the structures 
of PlaceNet below the last convolution layer are the same as ResNet50. The last 
convolution/pooling/fc layers are duplicated to five branches, namely, 
\emph{place}, \emph{category}, \emph{function}, \emph{city}, and 
\emph{country}, which is carefully designed for places. 
Each branch contains two FC layers.
Different loss 
functions and pooling methods are studied in this work.

\noindent\textbf{Loss Functions.} We study three losses for PlaceNet, namely, 
softmax loss, focal loss, and triplet loss. \emph{Softmax loss} or \emph{Focal loss} \cite{lin2017focal} is adopted to classify place, category, function, city, and country.
To learn the metric 
described by place pairs, we employ \emph{Triplet loss} \cite{schroff2015facenet}, which enforces distance constraints among positive and negative samples.
When using triplet loss, the network is optimized by a combination of $L_{softmax}$ and $L_{triplet}$ .

\noindent\textbf{Pooling Methods.} We also study different pooling methods for PlaceNet, 
namely, average pooling, max pooling, spatial pyramid pooling 
\cite{he2015spatial}.
Spatial pyramid pooling (SPP) is used to learn multi-scale pooling, which is 
robust to object deformations and can augment data to confront overfitting.


\begin{table}[t]
	\caption{The experimental results for different methods on all tasks. We vary different pooling methods and loss functions for PlaceNet.
	Except for the last line, models are trained on Places-Fine.
	The figures in bold/blue indicate optimal/sub-optimal performance, respectively}
	\label{tab:experiments}
\centering
\setlength{\tabcolsep}{2.5pt}
\scriptsize
{
\begin{tabular}{c|c|cc|cc|cc|cc|cc}
\hline
 \multicolumn{2}{c|}{} & \multicolumn{2}{c|}{Place} & 
 \multicolumn{2}{c|}{Category} & 
\multicolumn{2}{c|}{Function} & \multicolumn{2}{c|}{City} & 
\multicolumn{2}{c}{Country} \\
\hline
 \multicolumn{2}{c|}{} & Top-1 & Top-5 & Top-1 & Top-5 & Top-1 & 
Top-5 & Top-1 & Top-5 
& Top-1 & Top-5\\
\hline
\multirow{4}{*}{Backbone} & \multicolumn{1}{l|}{AlexNet} & 33.78 & 48.19 & 
24.16 & 53.03 & 64.97 
& 96.70 & 12.47 & 32.52 & 17.97 & 43.30\\
& \multicolumn{1}{l|}{GoogLeNet} & 53.48 & 66.23 & 26.01 & 54.81 & 
65.69 & 97.20 & 16.34 & 37.19 & 20.98 & 46.43\\
& \multicolumn{1}{l|}{VGG16} & 43.84 & 59.03 & 26.89 & \textcolor{blue}{55.68} & 65.97 
& 97.11 & 18.65 & \textcolor{blue}{41.13} & \textcolor{blue}{24.86} & 51.35\\
& \multicolumn{1}{l|}{ResNet50} & 54.53 & 67.01 & 25.22 & 53.62 & 
\textcolor{blue}{68.25} & 96.89 & 17.15 & 38.55 & 19.72 & 45.51\\
\hline
& \multicolumn{1}{l|}{Average} & 54.33 & \textcolor{blue}{67.66} & 25.95 & 
55.07 & 67.35 & 97.34 & \textcolor{blue}{18.73} & 40.30 & 24.80 & 51.03\\
Pooling & \multicolumn{1}{l|}{Max} & 49.66 & 63.26 & 25.11 & 54.07 & 
65.45 & 
 97.12 & 16.93 & 38.18 & 22.83 & 48.61\\
& \multicolumn{1}{l|}{SPP} & 28.18 & 45.55 & \textcolor{blue}{27.21} & 53.86 & 67.02 & 96.37 & 15.36 & 34.48 & 
21.00 & 43.08\\
\hline
& \multicolumn{1}{l|}{Softmax} & 54.31 & \textcolor{blue}{67.66} & 25.95 & 
55.07 & 67.35 & 97.34 & \textcolor{blue}{18.73} & 40.30 & 24.80 & 51.03\\
Loss & \multicolumn{1}{l|}{Triplet} & 50.33 & 64.06 & 21.15 & 48.92 & 
64.84 & 95.61 & 14.73 & 36.56 & 20.43 & 46.66\\
& \multicolumn{1}{l|}{Focal} & \textcolor{blue}{55.03} & 67.38 & 25.27 & 55.48 & 
67.62 & \textcolor{blue}{97.53} & 18.67 & 40.87 & 24.73 & \textcolor{blue}{51.46}\\
\hline
\multicolumn{2}{l|}{\tiny{PlaceNet on Places-Coarse}} & \textbf{67.85} & \textbf{79.35} & \textbf{40.42} & 
\textbf{68.98} & \textbf{75.48} & \textbf{97.58} & \textbf{29.25} & \textbf{53.47} & \textbf{35.83} & \textbf{63.78}\\
\hline
\end{tabular}

}
\end{table}

\begin{figure}[h]
	\center
	\includegraphics[width=0.9\linewidth]{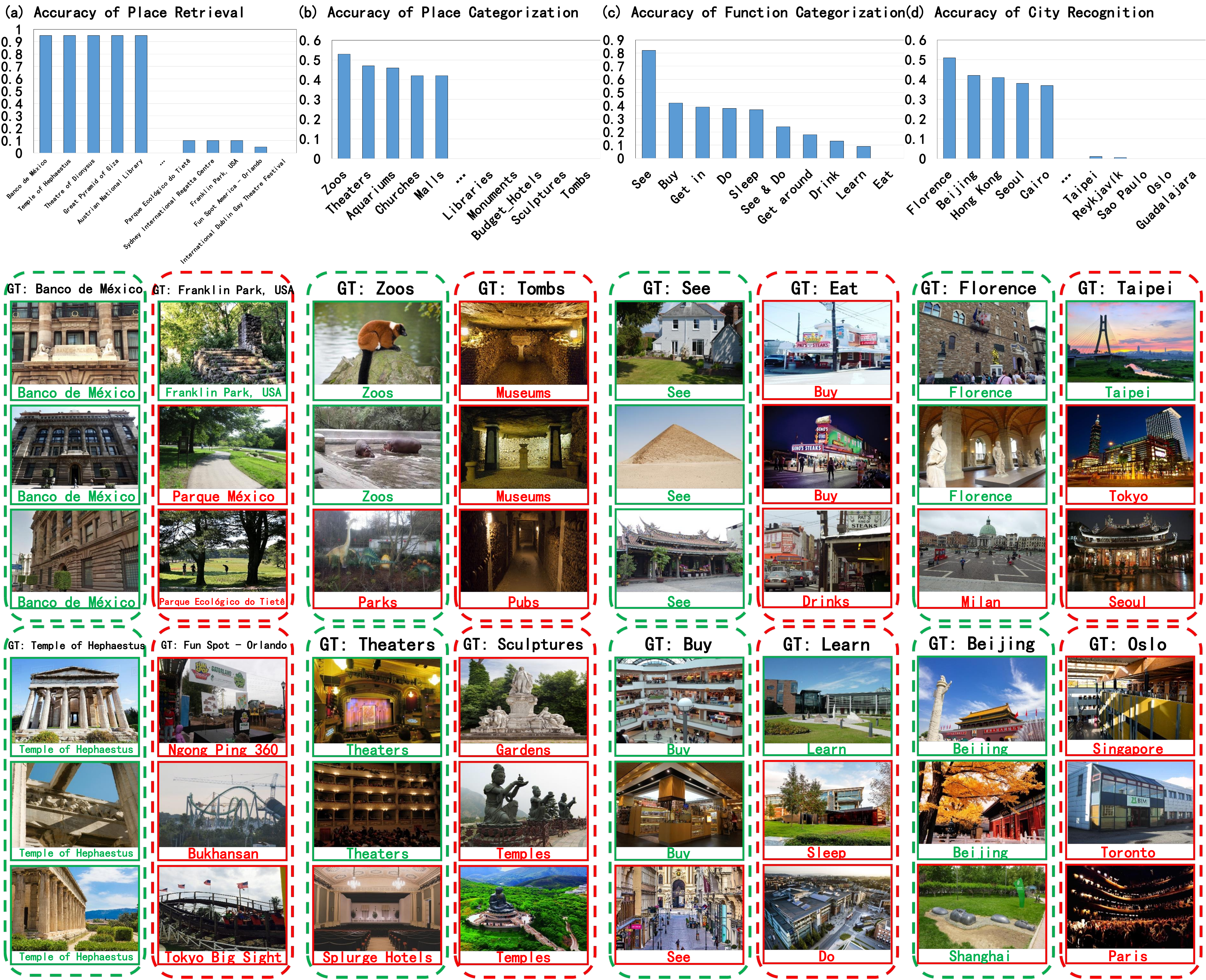}
	\caption{The $4$ tables show the performance of $4$ tasks, where each 
	presents the most and the least accurate $5$ classes. Below each table are 
	4 sets of examples, including 2 green/red dash boxes representing sample 
	classes with high/low accuracies. Inside each dash box is the ground truth 
	at the top and three images associated with predicted labels. Green/red 
	solid boxes of images mean right/wrong predictions}
	\label{fig_results}
\end{figure}

\subsection{Experimental Settings}
\noindent\textbf{Data.}
We use Places-Fine and Places-Coarse defined in Sec. \ref{benchmarks}
 as our experimental datasets. Note that Places-Fine and Places-Coarse share 
 the same 
 validation data and testing data, while training data size of the latter is much larger.

\noindent\textbf{Backbone Methods.}
Deep Convolutional Neural Networks (CNNs) \cite{he2015delving,hu2018squeeze,krizhevsky2012imagenet} have shown the impressive 
power for classification and retrieval tasks. 
Here we choose four popular CNN 
architectures, \textbf{AlexNet} \cite{krizhevsky2012imagenet}, 
\textbf{GoogLeNet} \cite{szegedy2015going}, \textbf{VGG16} 
\cite{simonyan2014very}, and 
\textbf{ResNet50} \cite{he2016deep}, then train them on Places-Fine to 
create backbone models.

\noindent\textbf{Training Details.} We train each model for $90$ epochs. For all tasks and all methods, the initial learning rate is set to be $0.5$. And the learning rate is multiplied by $0.1$ at epoch $63$ and epoch $81$. The weight decay is $1e^{-4}$. For the optimizer, we use stochastic gradient descent with $0.9$ momentum. We also augment the data following the operation on ImageNet, including randomly cropping and horizontally flipping the images.
All images are resized to $224\times224$ and normalized with mean $[123,117,109]$ and standard deviation $[58,56,58]$.
Each model is pre-trained on ImageNet and then trained with our Placepedia Dataset in an end-to-end manner.

All experiments are conducted on Places-Fine. And we also
train our PlaceNet on Places-Coarse to see if larger scale of datasets can 
further
benefit the recognition performance.

\subsection{Analysis on Recognition Results}
Quantitative evaluations of different methods on the four 
benchmarks are provided.
Table \ref{tab:experiments} summarizes the performance 
of different methods on all tasks. We first analyze the results on Places-Fine for all benchmark tasks.

\noindent\textbf{Place Retrieval.}
PlaceNet with focal loss achieves the best 
retrieval results 
when evaluated using the top-$1$ 
accuracy. Some sample places with high/low accuracies are shown in Fig. 
\ref{fig_results} (a). We observe that: 1) Places with distinctive architectures  
can be easily recognized, \eg \emph{Banco de M\'exico} and \emph{Temple of Hephaestus}. 2) For some parks, \eg \emph{Franklin Park}, there is 
usually no clear evidence to tell them from other parks. The same scenario can 
take place in categories such as gardens and churches. 3) Big 
places like \emph{Fun Spot Orlando} may contain several small places, where their appearance may have nothing in common, which makes it very difficult to 
recognize. Places like resorts, towns, parks, and universities suffer the same 
issue.

\noindent\textbf{Place Categorization.}
The best result is yielded by PlaceNet plus SPP.
Some sample categories with high/low 
accuracies are shown in Fig. \ref{fig_results} (b). We observe that: 1) \emph{Zoos} are the most distinctive. Intuitively, if animals are seen in a place, 
that is probably a zoo. However, photos in zoos may be mistaken for taking in 
parks. 2) \emph{Tombs} can be confused with \emph{Pubs}, due to bad 
illumination condition.

\noindent\textbf{Function Categorization.}
The best setting for learning the function of a place is to use ResNet models. Some sample functions with high/low accuracies are shown in 
Fig. \ref{fig_results} (c). 1) \emph{See} is recognized with the highest accuracy.
2) Some examples of \emph{Buy} are very difficult to identify, \eg the third 
image in \emph{Buy}. Even human cannot tell what a street is mainly used for. 
Is it for shopping, eating, or just for transportation?
Same logic applies to shops. The images of \emph{Eat} are often 
categorized as shops for \emph{buy}ing or \emph{drink}ing. One possible way to recognize the 
function of a shop is to extract and analyze its name, or to recognize and 
classify the food type therein.
3) Universities are often unrecognized either, due to its large area with 
various buildings/scenes.

\noindent\textbf{City/Country Recognition.}
From Fig. \ref{fig_results} (d), we observe that:
1) Cities with long history (\eg \emph{Florence}, \emph{Beijing}, \emph{Cairo}) are more likely to 
distinguish from others, because they often preserve the oldest arts and 
architectures.
2) Travelers often conclude that Taiwan and Japan look 
quite alike. The results do show that places of \emph{Taipei} may be regarded as in \emph{Tokyo}.
3) Although places can be wrongly classified to another city, the prediction often belongs to the same country with the ground truth city. For 
instance, Florence and Milan are both in Italy; Beijing and Shanghai are both 
in China.
The results of country recognition are not presented here. They demonstrate similar findings to city recognition.

To conclude, we see that place-related tasks are often very challenging:
1) Places of parks, gardens, churches, \etc are easy to classify; However, it is difficult to distinguish one park/garden/church from another.
2) Under bad environmental condition, photos can be extremely difficult to categorize;
3) To recognize the function of a street or a shop is non-trivial, \ie it is hard to determine their use for people to have a dinner, take a drink, or go shopping.
4) Cities of long history such as Beijing and Florence are often recognized with a high accuracy. While images of others are more likely to be misclassified as similar ones inside and outside their countries.
We hope that Placepedia with its well-defined benchmarks can foster more effective studies and thus benefit place recognition and retrieval.
The last line of Tab. \ref{tab:experiments} shows that, to train PlaceNet on 
larger amount of data, we can further obtain performance gain, by $7$ to 
$16$ percent on different tasks.

\section{Study on Multi-Faceted City Embedding}

We embed each city in an expressive vector to understand places on a city level. Also, the connections between the visual characteristics and the underlying economic or cultural implications are studied therefrom.

\subsection{City Embedding}
\label{embedding_method}
City embedding is to use a vector to represent a city, the items of 
which indicate different aspects, \eg the economy level, the cultural deposits, the politics atmosphere, \etc. In this study, cities are embedded from  
both visual and textual domains.
1) Visual representations of cities are obtained by extracting features from models supervised by city items.
2) Leading paragraphs collected from Wikipedia are used as the description of each city.
\cite{devlin2018bert} provided a pre-trained model on language understanding to 
embed the content of texts into numeric space. We use this model to extract the 
textual representations for all cities.

\noindent\textbf{Network Structure.} The model for city embedding is illustrated in 
Figure \ref{fig_networks} (b). The input is constructed by concatenating visual and 
textual vectors. Two fully connected layers are then applied to learn city embedding 
representations. The corresponding activation functions are ReLU. At last, a classifier and cross entropy loss are used to supervise the learning procedure.

\noindent\textbf{Representative Vectors.} We train the network iteratively. The 
well-trained network is then used to extract the embedding vectors for all 
images. City embeddings are then acquired by averaging image embeddings city-wise.

\subsection{Experimental Results}
We analyze city embedding results from two aspects. Firstly, we compare the expressive power of embeddings using different information, namely vision, text, and vision \& text, in order to see if learning from both can yield a better city representation. Secondly, we investigate the embedding results neuron-wise to explore what kinds of images can express economic/cultural/political levels of cities the most.

\begin{figure}[t]
	\center
	\includegraphics[width=0.9\linewidth]{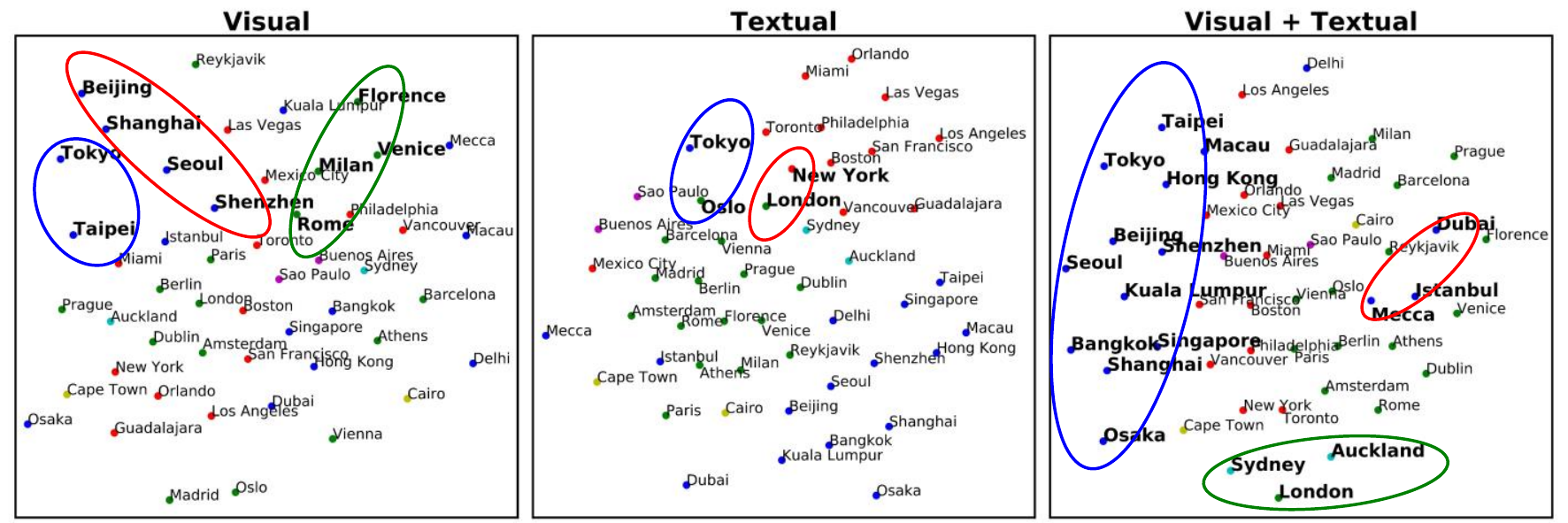}
	\caption{These three figures show t-SNE representation for city embeddings using vision, text, and vision \& text info, respectively. Points with the same color belong to the same continent. We can see that learning from both generates the best embedding results}
	\label{fig_embedding}
\end{figure}

\noindent\textbf{Visual and Text Embedding.}
In Fig. \ref{fig_embedding}, we demonstrate three embedding results using t-SNE \cite{maaten2008visualizing}.
1) The left graph shows embeddings using only visual features
. We observe that it tends to cluster cities that are visually similar. For example, Tokyo looks like Taipei; Beijing, Shanghai, and Shenzhen are all from China, and Seoul shares lots of similar architectures with them; Florence, Venice, Milan, and Rome are all Italy cities.
2) The graph on the middle shows embeddings using only textual features.
We can see that textual features usually express the functionality and geolocation of a city. For example, Tokyo and Oslo are both capitals; London and New York are both financial centers. However, they are not visually alike. Also, cities from the same continent are clustered.
3) The right graph shows embeddings learned from both visual and textual domains. They can express the resemblance visually and functionally. For example, cities from east/west-Asia are all clustered together, and cities from Commonwealth of Nations like Sydney, Auckland, and London, are also close to each other on the graph.
From the comparison of these graphs, we conclude that learning embeddings from both vision and text content produces the most expressive power of cities.

\begin{figure}[t]
	\center
	\includegraphics[width=0.5\linewidth]{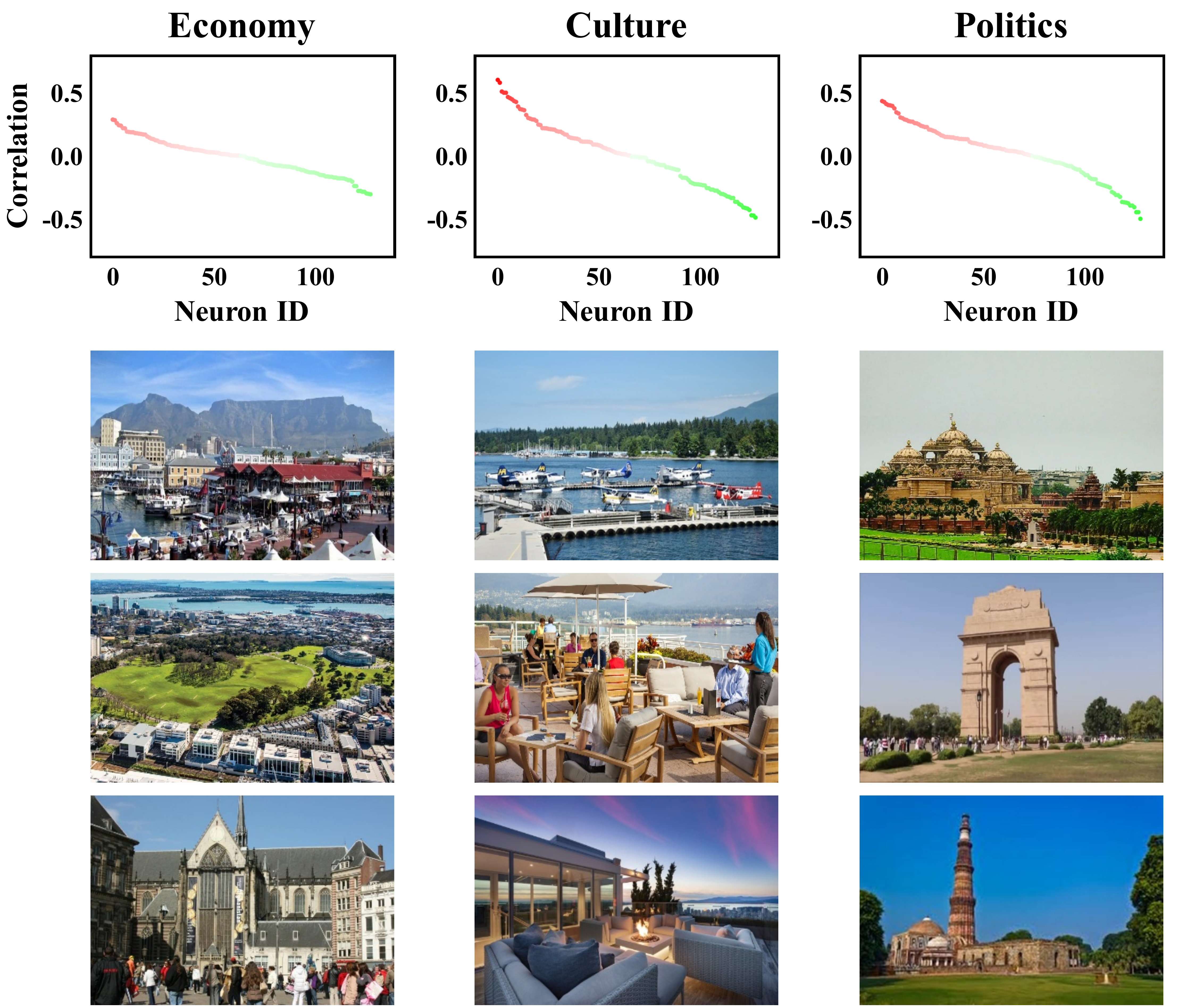}
	\caption{Three graphs rank Pearson correlation based on neurons in terms of \emph{economy}, \emph{culture}, and \emph{politics}. Below each presents top-$3$ places activating the neuron of the largest correlation value}
	\label{fig_ecp}
\end{figure}

\noindent\textbf{Economic, Cultural, or Political.}
We follow the work in \cite{son2018your} to represent each city in three dimensions, namely \emph{economy}, \emph{culture}, and \emph{politics}. In \cite{son2018your}, word lists indicating \emph{economy}, \emph{culture}, \emph{politics} are predefined. In this work, leading paragraphs of Wikipedia pages are adopted as our city description. For each city, we calculate the weights of \emph{economic}, \emph{cultural}, and \emph{political} therefrom as in \cite{son2018your}. And we match each neuron to them using Pearson correlation \cite{benesty2009pearson}, in order to quantify the connection between each neuron and them. Quinnipiac University\footnote{http://faculty.quinnipiac.edu/ libarts/polsci/Statistics.html} concludes that a correlation above $0.4$ or below $-0.4$ can be viewed as a strong correlation. From Fig. \ref{fig_ecp}, we see that neurons can express \emph{culture} most confidently, with the highest correlation score larger than $0.6$. This is consistent with our knowledge, \ie culture usually is expressed from distinctive architectures or some unique human activities.
Looking at the top-$3$ places that activate the most relevant neuron, we observe that: 1) Economy level is usually conveyed by a cluster of buildings or the crowd on streets, indicating a prosperous place; 2) Cultural atmosphere can be expressed by distinguished architecture styles and human activities; (3) Political elements are often related to temples, churches, and some historical sites, which usually indicate religious activities and politics-related historical movements.



\section{Conclusion}
In this work, we construct a large-scale place dataset which is comprehensively annotated with multiple aspects. To our knowledge, it is the largest place-related dataset available. To explore place understanding, we carefully build several benchmarks and study contemporary models. The experimental results show that there still remains lots of challenges in place recognition. To learn city embedding representations, we demonstrate that learning from both visual and textual domains can better characterize a city. The learned embeddings also demonstrate that \emph{economic}, \emph{cultural}, and \emph{political} elements can be represented in different types of images. We hope that, with comprehensively annotated Placepedia contributed to the community, more powerful and robust systems will be developed to foster future place-related studies.

\noindent \textbf{Acknowledgment}
This work is partially supported by the SenseTime Collaborative Grant 
on 
Large-scale Multi-modality Analysis (CUHK Agreement No. TS1610626 \& No. 
TS1712093), the General Research Fund (GRF) of Hong Kong (No. 14203518 \& No. 
14205719).


\newpage
\section*{Supplementary Material}

\subsection*{Annotation Interface}
Fig. \ref{fig_annotation} shows the interface of the annotation tool.




\subsection*{Additional Results}

\begin{figure*}[h]
 \begin{minipage}[b]{0.49\textwidth}
  \centering
	\includegraphics[width=\linewidth]{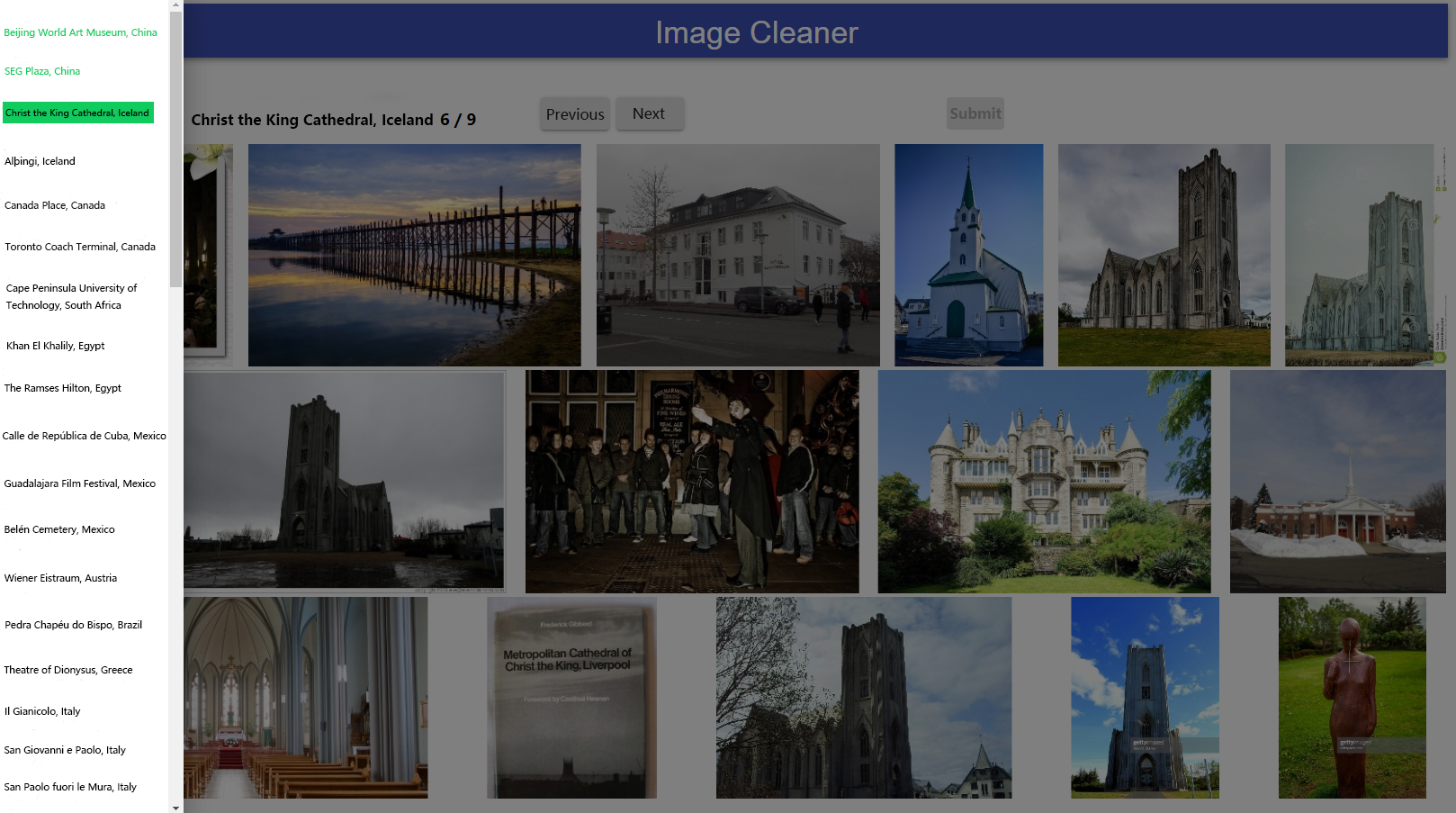}
	\caption{
		Interface of the annotation tool. The left column lists all places. Text in green means that images of the place have been annotated; Text with green background represents the current place to clean; Text in black indicates places waiting to be cleaned
	}
	\label{fig_annotation}
 \end{minipage}
 \hfill
 \begin{minipage}[b]{0.49\textwidth}
  \centering
	\includegraphics[width=0.61\linewidth]{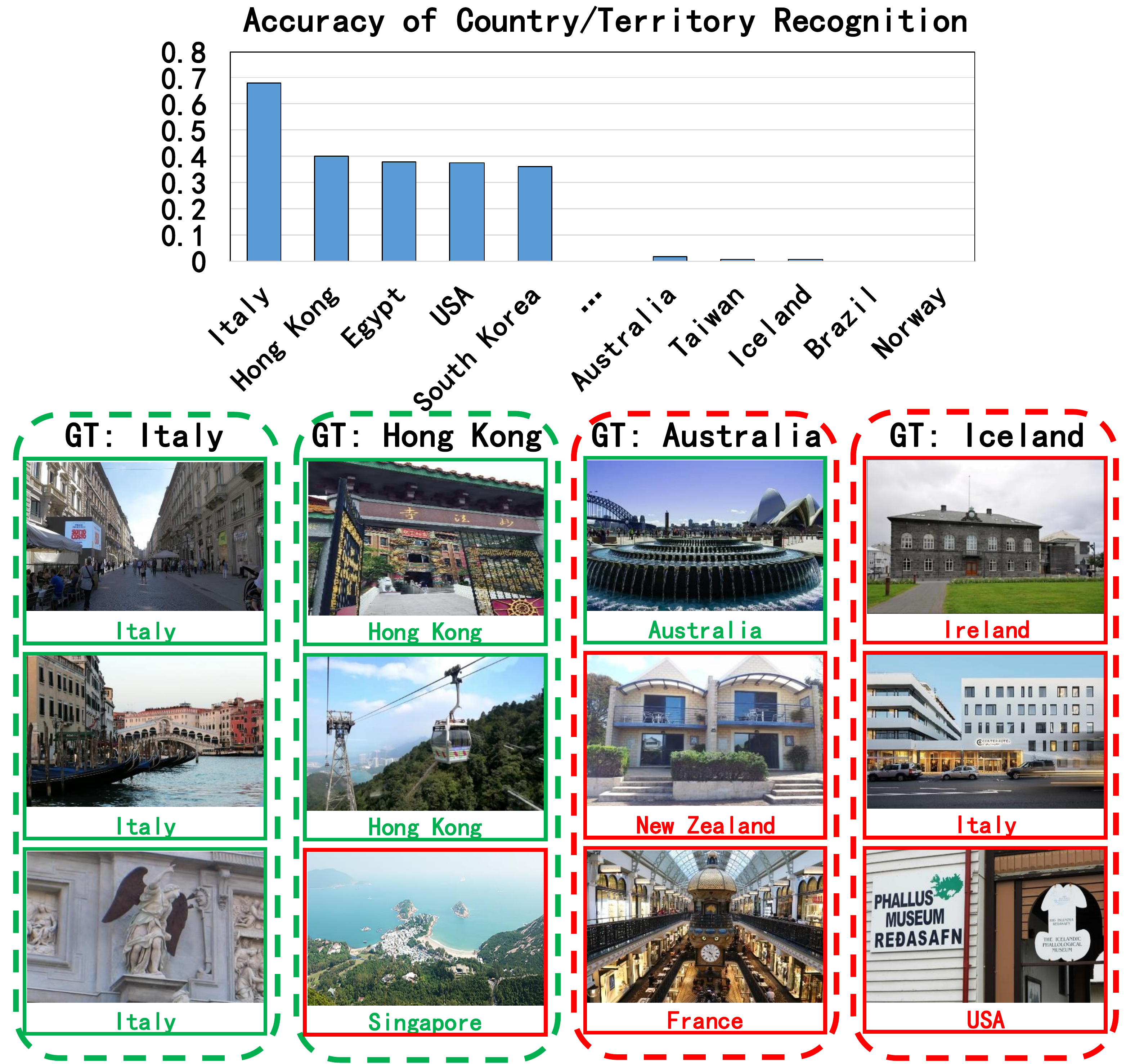}
	\caption{
		The table shows the most and the least accurate 5 classes of country recognition. Below the table are 4 sets of examples, including 2 green/red dash boxes representing sample classes with high/low accuracies. Inside each dash box is the ground truth at the top and three images associated with predicted labels. Green/red solid boxes of images mean right/wrong predictions
	}
	\label{fig_country}
 \end{minipage}
\end{figure*}

\paragraph{Country Recognition}
Fig. \ref{fig_country} shows the results of country recognition. They reveal similar findings to city recognition. Although places can be wrongly classified to another country, the prediction often belongs to the same continent with the ground truth country. For instance, Australia and New Zealand are both in Oceania; Iceland, Ireland, and Italy are all in Europe.

\paragraph{Weights of Economy, Culture, and Politics}
Tab. \ref{tab:weights} shows the weights of economy, culture, and politics for all cities of Places-Fine, which are computed as in \cite{son2018your}. Lists of pre-defined economic/cultural/political keywords are provided therein. We use the leading paragraphs of Wikipedia pages as city description, and calculate the frequencies of these three types of keywords. By normalizing the frequencies we get the weights of economy, culture, and politics for each city.

\begin{table*}[]
    \centering
    \scriptsize
\caption{The weights of economy, culture, and politics of 50 cities of 
Places-Fine, which are computed as in \cite{son2018your}, using the leading 
paragraphs of Wikipedia pages as city description}
    \begin{tabular}{l|l|l|l|l|l}
    \hline
    & & & Economic & Cultural & Political \\
    City               & Country / Territory      & Continent     & keywords & keywords & keywords \\ \hline
    Amsterdam          & Netherlands              & Europe        & 0.331             & 0.294             & 0.375              \\ \hline
    Athens             & Greece                   & Europe        & 0.299             & 0.278             & 0.424              \\ \hline
    Auckland           & New Zealand              & Oceania       & 0.320             & 0.310             & 0.370              \\ \hline
    Bangkok            & Thailand                 & Asia          & 0.341             & 0.307             & 0.352              \\ \hline
    Barcelona          & Spain                    & Europe        & 0.336             & 0.336             & 0.328              \\ \hline
    Beijing            & China                    & Asia          & 0.345             & 0.310             & 0.345              \\ \hline
    Berlin             & Germany                  & Europe        & 0.323             & 0.263             & 0.414              \\ \hline
    Boston             & United States of America & North America & 0.317             & 0.292             & 0.392              \\ \hline
    Buenos Aires       & Argentina                & South America & 0.310             & 0.296             & 0.394              \\ \hline
    Cairo              & Egypt                    & Africa        & 0.333             & 0.282             & 0.385              \\ \hline
    Cape Town          & South Africa             & Africa        & 0.358             & 0.245             & 0.396              \\ \hline
    Delhi              & India                    & Asia          & 0.320             & 0.262             & 0.418              \\ \hline
    Dubai              & United Arab Emirates     & Asia          & 0.426             & 0.259             & 0.315              \\ \hline
    Dublin             & Ireland                  & Europe        & 0.320             & 0.280             & 0.400              \\ \hline
    Florence           & Italy                    & Europe        & 0.323             & 0.338             & 0.338              \\ \hline
    Guadalajara        & Mexico                   & North America & 0.394             & 0.282             & 0.324              \\ \hline
    Hong Kong          & Hong Kong                    & Asia      & 0.315             & 0.204             & 0.481              \\ \hline
    Istanbul           & Turkey                   & Europe / Asia & 0.317             & 0.275             & 0.408              \\ \hline
    Kuala Lumpur       & Malaysia                 & Asia          & 0.272             & 0.272             & 0.456              \\ \hline
    Las Vegas          & United States of America & North America & 0.319             & 0.319             & 0.362              \\ \hline
    London             & United Kingdom           & Europe        & 0.335             & 0.335             & 0.330              \\ \hline
    Los Angeles      & United States of America & North America & 0.326             & 0.302             & 0.372              \\ \hline
    Macau              & Macau                    & Asia          & 0.349             & 0.233             & 0.419              \\ \hline
    Madrid             & Spain                    & Europe        & 0.286             & 0.299             & 0.416              \\ \hline
    Mecca              & Saudi Arabia             & Asia          & 0.288             & 0.288             & 0.425              \\ \hline
    Mexico City        & Mexico                   & North America & 0.308             & 0.260             & 0.432              \\ \hline
    Miami              & United States of America & North America & 0.360             & 0.315             & 0.326              \\ \hline
    Milan              & Italy                    & Europe        & 0.341             & 0.287             & 0.372              \\ \hline
    New York City      & United States of America & North America & 0.328             & 0.323             & 0.349              \\ \hline
    Orlando            & United States of America & North America & 0.313             & 0.297             & 0.391              \\ \hline
    Osaka              & Japan                    & Asia          & 0.303             & 0.242             & 0.455              \\ \hline
    Oslo               & Norway                   & Europe        & 0.400             & 0.247             & 0.353              \\ \hline
    Paris              & France                   & Europe        & 0.323             & 0.311             & 0.366              \\ \hline
    Philadelphia       & United States of America & North America & 0.325             & 0.299             & 0.376              \\ \hline
    Prague             & Czech Republic           & Europe        & 0.326             & 0.304             & 0.370              \\ \hline
    Reykjavík          & Iceland                  & Europe        & 0.321             & 0.286             & 0.393              \\ \hline
    Rome               & Italy                    & Europe        & 0.328             & 0.259             & 0.414              \\ \hline
    San Francisco      & United States of America & North America & 0.322             & 0.308             & 0.370              \\ \hline
    São Paulo          & Brazil                   & North America & 0.333             & 0.306             & 0.361              \\ \hline
    Seoul              & South Korea              & Asia          & 0.341             & 0.302             & 0.357              \\ \hline
    Shanghai           & China                    & Asia          & 0.373             & 0.273             & 0.355              \\ \hline
    Shenzhen           & China                    & Asia          & 0.339             & 0.303             & 0.358              \\ \hline
    Singapore          & Singapore                & Asia          & 0.311             & 0.316             & 0.373              \\ \hline
    Sydney             & Australia                & Oceania       & 0.333             & 0.311             & 0.356              \\ \hline
    Taipei             & Taiwan                    & Asia          & 0.318             & 0.291             & 0.391              \\ \hline
    Tokyo              & Japan                    & Asia          & 0.317             & 0.289             & 0.394              \\ \hline
    Toronto            & Canada                   & North America & 0.340             & 0.321             & 0.340              \\ \hline
    Vancouver          & Canada                   & North America & 0.315             & 0.324             & 0.360              \\ \hline
    Venice             & Italy                    & Europe        & 0.306             & 0.296             & 0.398              \\ \hline
    Vienna             & Austria                  & Europe        & 0.306             & 0.298             & 0.397              \\ \hline
    \end{tabular}
    \label{tab:weights}
\end{table*}

\end{document}